\documentclass{article}
\usepackage{spconf,amsmath,epsfig}
\usepackage{scalefnt}
\usepackage{url}

\title{Hue Histograms to Spatiotemporal Local Features \\for Action Recognition}
\name{Fillipe Souza$^{(1)}$, Eduardo Valle$^{(2)}$, Guillermo Ch\'avez$^{(3)}$, Arnaldo Ara\'ujo$^{(1)}$\thanks{We would like to thank the FAPEMIG, CNPq, FAPESP and Capes agencies for the financial support.}}
\address{(1) NPDI Lab -- DCC/UFMG -- Belo Horizonte/MG, Brazil -- \{fdms,arnaldo\}@dcc.ufmg.br\\
	(2) RECOD Lab -- IC/Unicamp -- Campinas/SP, Brazil -- mail@eduardovalle.com\\
	 (3) ICEB/UFOP -- Ouro Preto/MG, Brazil -- gcamarac@gmail.com\\}
\begin{document}
%\ninept
%
\maketitle
\begin{abstract}
Despite the recent developments in spatiotemporal local features for action recognition in 
video sequences, local color information has so far been ignored. However, 
color has been proved an important element to the success of automated recognition 
of objects and scenes. In this paper we extend the space-time interest point descriptor STIP 
to take into account the color information on the features' neighborhood. We compare 
the performance of our color-aware version of STIP (which we have called HueSTIP) 
with the original one.
\end{abstract}
\section{Introduction}
\label{sec:intro}

In this work we provide a discussion on the role of spatiotemporal color features for human 
action recognition in realistic settings. Color is a prominent feature of the real world scenes 
and objects. Not surprisingly, it has become a powerful tool in automated object 
recognition~\cite{gevers04}~\cite{weijer06}~\cite{vande10}. However, color has not yet been 
given its deserved importance in the universe of unconstrained action recognition. 

Several spatiotemporal local feature descriptors and detectors have been proposed and evaluated 
in action recognition. Detectors rely commonly on a measure function (or response function) to 
locate interest regions. Those local regions (also called patches) can be described in terms, 
for example, of histograms of gradient orientations and optical flow. Laptev~\cite{laptev05} 
presented a spatiotemporal extension of the Harris-Laplace corner detector proposed by Mikolajczyk 
and Schmid~\cite{ms04}. Spatiotemporal corners are found when strong intensity variations over the 
spatial and temporal domains occur simultaneously. This method was proved efficient for action 
recognition in controlled datasets such as the KTH dataset~\cite{WUKLS09}. Doll\'ar et al.
\cite{dollar05} proposed a spatiotemporal detector based on temporal Gabor filters that considers 
only local variations having periodic frequency components. Another spatiotemporal interest point 
detector was designed in~\cite{willems08} by Willems et al. This detector uses the Hessian 
determinant as a saliency measure and 3D convolution approximations by box-filters %~\cite{bay07}~\cite{ke05} 
in order to find regions of interest. Here we will only provide a formal discussion 
of the spatiotemporal local feature detector used in this work, the one proposed in~\cite{laptev05}.

To improve the discriminative power and illumination invariance of local features to object recognition 
and image categorization, a set of color descriptors for spatial local features was proposed in~\cite{vande10} 
by Sande et al. for static images. The distictiveness of their color descriptors was evaluated 
experimentally and their invariant properties under illuminations changes were analyzed. They derived different 
color descriptors, including combinations with the intensity-based shape descriptor SIFT~\cite{lowe04}. 
Weijer et al.~\cite{weijer06} had already proposed color histograms providing robustness to 
photometric and geometrical changes, photometric stability and generality. Their work was the basis for 
some of the descriptors developed in~\cite{vande10}, also for our own.

The most important contribution of this work is the combination of the work Weijer et al.~\cite{weijer06} 
and Laptev~\cite{laptev05} to propose a very discriminating and robust local descriptor for videos, which takes 
into account the color information on the features' neighborhood. The second contribution is the evaluation of 
this descriptor and its comparison with STIP in the challenging database Hollywood2 of Marszalek et al.~\cite{marszalek09}, 
with a detailed analysis of the cases of success and failure brought by the addition of color information.

The rest of this paper is organized as follows. In section~\ref{sec:stip} we are concerned with the formal 
description of the spatiotemporal interest point detector used. Further, the details on the color descriptors 
are presented in section~\ref{sec:colordescriptor}. The experiments and their results are discussed in 
section~\ref{sec:experiments} and section~\ref{sec:conclusion} concludes the work.

%\begin{table}[htb]
%\begin{center}
%\caption{A summary of the test database for pornography detection.}\label{tab: tab1}
%\begin{tabular}{llll}
%\hline
%\textbf{Class} & \multicolumn{1}{|c|}{\textbf{Videos}} & \multicolumn{1}{c|}{\textbf{Hours}} & \multicolumn{1}{c}{\textbf{Shots per Video}} \\
%\hline
%Porn & \multicolumn{1}{|c|}{400} & \multicolumn{1}{c|}{57} & \multicolumn{1}{c}{15.6} \\
%Non-porn (``Easy'') & \multicolumn{1}{|c|}{200} & \multicolumn{1}{c|}{11.5} & \multicolumn{1}{c}{33.8} \\
%Non-porn (``Difficult'') & \multicolumn{1}{|c|}{200} & \multicolumn{1}{c|}{8.5} & \multicolumn{1}{c}{17.5} \\
%\hline
%\textbf{All videos} & \multicolumn{1}{|c|}{\textbf{800}} & \multicolumn{1}{c|}{\textbf{77}} & \multicolumn{1}{c}{\textbf{20.6}} \\
%\hline
%\end{tabular}
%\end{center}
%\end{table}

\section{Spatiotemporal Interest Points}
\label{sec:stip}

Laptev~\cite{laptev05} designed a differential operator that checks for extremas 
over the spatial and temporal scales. Those extremas in specific space-time locations 
refer to particular patterns of events. This method is built on the Harris~\cite{bb22952} 
and F\"orstner~\cite{bb22653} interest point operators, but extended to the temporal space. 
Essentially, as a corner moves across an image sequence, at the change of its direction 
an interest point is identified. Other typical situations are when image structures are 
either split or unified. For being one of the major elements in this work, a few details and 
mathematical considerations on the detector design are presented next.

Many interest events in videos are characterized by motion variations of 
image structures over time. In order to retain those important information, the 
concept of spatial interest points is extended to the spatio-temporal domain. 
This way, the local regions around the interest points are described with respect 
to derivatives in both directions (space and time).

At first, the selection of interest point in the spatial domain is described.
The linear scale-space representation of an image can be mathematically defined 
as $L^{sp}:R^2 \times R_+ \mapsto R$, which is the convolution of 
$f^{sp}$ with $g^{sp}$, where $f^{sp}:R^2 \mapsto R$ represents a simple model 
of an image and $g^{sp}$ is the Gaussian kernel of variance $\sigma_{l}^2$. Then, 

\begin{equation}
L^{sp}(x,y;\sigma_{l}^2)=g^{sp}(x,y;\sigma_{l}^2) \ast f^{sp}(x,y),
\end{equation} and

\begin{equation}
g^{sp}(x,y;\sigma_{l}^2)=\frac{1}{2\pi\sigma_{l}^2}\exp(-(x^2+y^2)/2\sigma_{l}^2).
\end{equation}

Localizing interest points means to find strong variations of image intensities 
along the two directions of the image. To determine those local regions, the 
second moment matrix is integrated over a Gaussian window having variance 
$\sigma_{i}^2$, for different scales of observation $\sigma_{l}^2$, which 
is written as the equation:

\begin{eqnarray}
\mu^{sp}(.;\sigma_{l}^2, \sigma_{i}^2)&=&g^{sp}(.;\sigma_{i}^2) \ast ((\nabla L(.;\sigma_{l}^2))(\nabla L(.;\sigma_{l}^2))^T)\nonumber\\
&=&g^{sp}(.;\sigma_{i}^2) \ast \left(\begin{array}{rr}
(L_x^{sp})^2&L_x^{sp}L_y^{sp}\\
L_x^{sp}L_y^{sp}&(L_y^{sp})^2.
\end{array}\right)
\label{eq:sp}
\end{eqnarray}

The descriptors of variations along the dimensions of $f^{sp}$ are the eigenvalues 
of Equation~\ref{eq:sp}: $\lambda_1$ and $\lambda_2$, with $\lambda_1 \leq \lambda_2$. 
Higher values of those eigenvalues is a sign of interest point and generally 
leads to positive local maxima of the Harris corner function, provided that 
the ratio $\alpha=\lambda_2/\lambda_1$ is high and satisfies the constraint 
$k \leq \alpha/(1+\alpha)^2$:

\begin{eqnarray}
H^{sp}&=&\det(\mu^{sp})-k.trace^2(\mu^{sp})\nonumber\\
&=&\lambda_1\lambda_2-k(\lambda_1+\lambda_2)^2
\end{eqnarray}.

Analogously, the procedure to detect interest points in the scape-time domain is 
derived by rewriting the equations to consider the temporal dimension. Thus,  
having an image sequence modeled as $f:R^2 \times R \mapsto R$, its linear 
representation becomes $L:R^2 \times R \times R_+^2 \mapsto R$, but over two 
independent variances $\sigma_{l}^2$ (spatial) and $\tau_{l}^2$ (temporal) using 
an anisotropic Gaussian kernel $g(.;\sigma_{l}^2, \tau_{l}^2)$. Therefore, the 
complete set of equations for detecting interest points described in~\cite{laptev05} 
is the following.

\begin{equation}
L(.;\sigma_{l}^2)=g(.;\sigma_{l}^2, \tau_{l}^2) \ast f(.),
\end{equation}

\begin{eqnarray}
g(x,y,t;\sigma_{l}^2, \tau_{l}^2)=\frac{1}{\sqrt{(2\pi)^3\sigma_{l}^4\tau_{l}^2}}\nonumber\\
\times\exp(-(x^2+y^2)/2\sigma_{l}^2-t^2/\tau_{l}^2),
\end{eqnarray}

\begin{eqnarray}
\mu=g(.;\sigma_{i}^2) \ast \left(\begin{array}{rrr}
L_x^2&L_xL_y&L_xL_t\\
L_xL_y&L_y^2&L_yL_t\\
L_xL_t&L_yL_t&L_t^2.
\end{array}\right)
\label{eq:t}
\end{eqnarray}

\begin{eqnarray}
H&=&\det(\mu)-k.trace^3(\mu)\nonumber\\
&=&\lambda_1\lambda_2\lambda_3-k(\lambda_1+\lambda_2+\lambda_3)^3,
\end{eqnarray}
restricted to $H \geq 0$, with $\alpha = \lambda_2/\lambda_1$ and $\beta = 
\lambda_3/\lambda_1$, and subject to $k \leq \alpha\beta/(1+\alpha+\beta)^3$ 

\section{Local Features}
\label{sec:colordescriptor}

Given a local interest region denoted by a spatiotemporal interest point $(x,y,t,\sigma,\tau)$, 
3D local features accounting for appearance (histograms of oriented gradient) and motion (histograms 
of optical flow) are computed by using information from the neighborhood at $(x,y,t)$.
A spatiotemporal volume is sliced into $n_x\times n_y\times n_t$ 3D cells, in particular, $n_x=n_y=3$ 
and $n_t=2$. For each cell 4-bin histograms of gradient orientations (\textbf{HoG}) and 5-bin histograms of 
optical flow (\textbf{HoF}) are calculated, normalized and concatenated (\textbf{HoGHoF}, used 
by STIP~\cite{laptev05}).

\subsection{Color Descriptor}

In this section, the hue histogram based color descriptor is roughly described. From the work 
in~\cite{weijer06}, the hue calculation has the form:

\begin{equation}
hue=\arctan{(\frac{\sqrt{3}(R-G)}{R+G-2B})}.
\label{eq:hue}
\end{equation}

It is known that, in the HSV %(\textbf{H}ue, \textbf{S}aturation, \textbf{V}alue) 
color space, the hue value becomes unstable as it approaches the grey axis. 
In attempt to atenuate this problem, Weijer et al.~\cite{weijer06} analyzed the error propagation in 
the hue transformation and verified the inverse proportionality of the hue certainty to the saturation. 
This way, the authors demonstrated that the hue color model achieves robustness by weighing the hue sample by 
the corresponding saturation, which is given by Equation~\ref{eq:sat}:

\begin{equation}
sat=\sqrt{\frac{2(R^2 + G^2 + B^2 - RG - RB - GB)}{3}}.
\label{eq:sat}
\end{equation}

To construct the hue histogram, we calculate the bin number to which the hue value (of the spatiotemporal 
volume) belongs with $bin=hue * 36 / 2\pi$. Then, at the position $bin$ of the histogram the saturation value 
is accumulated. Before incrementing the histogram bin with a given amount of saturation, the saturation is 
weighed by a corresponding value of a Gaussian mask having the size of the spatiotemporal volume. The size and 
values forming the spatiotemporal Gaussian mask will vary according to the spatial and temporal scales of 
the interest point. The computed hue histogram will be further concatenated to the HoGHoF feature vector and 
this combination will be called \textbf{HueSTIP}. 

%\subsection{Another Hue Histogram (o que seria um bom título? vamos colocar este resultados?)}

%\begin{equation}
%hue=\arctan{(\frac{\sqrt{3}(R-G)}{R+G-2B})}. 
%\end{equation}

%\begin{equation}

%sat=
%\end{equation}

\section{Experiments}
\label{sec:experiments}

In our experiments, we investigated the power of the spatiotemporal local features containing color information for action recognition. This section describes the experimental setup followed by the analysis of the obtained results. 

\subsection{Dataset}
\label{sec:dataset}

\begin{figure}[htb]

%\begin{minipage}[b]{1.0\linewidth}
  \centering
 \centerline{\epsfig{figure=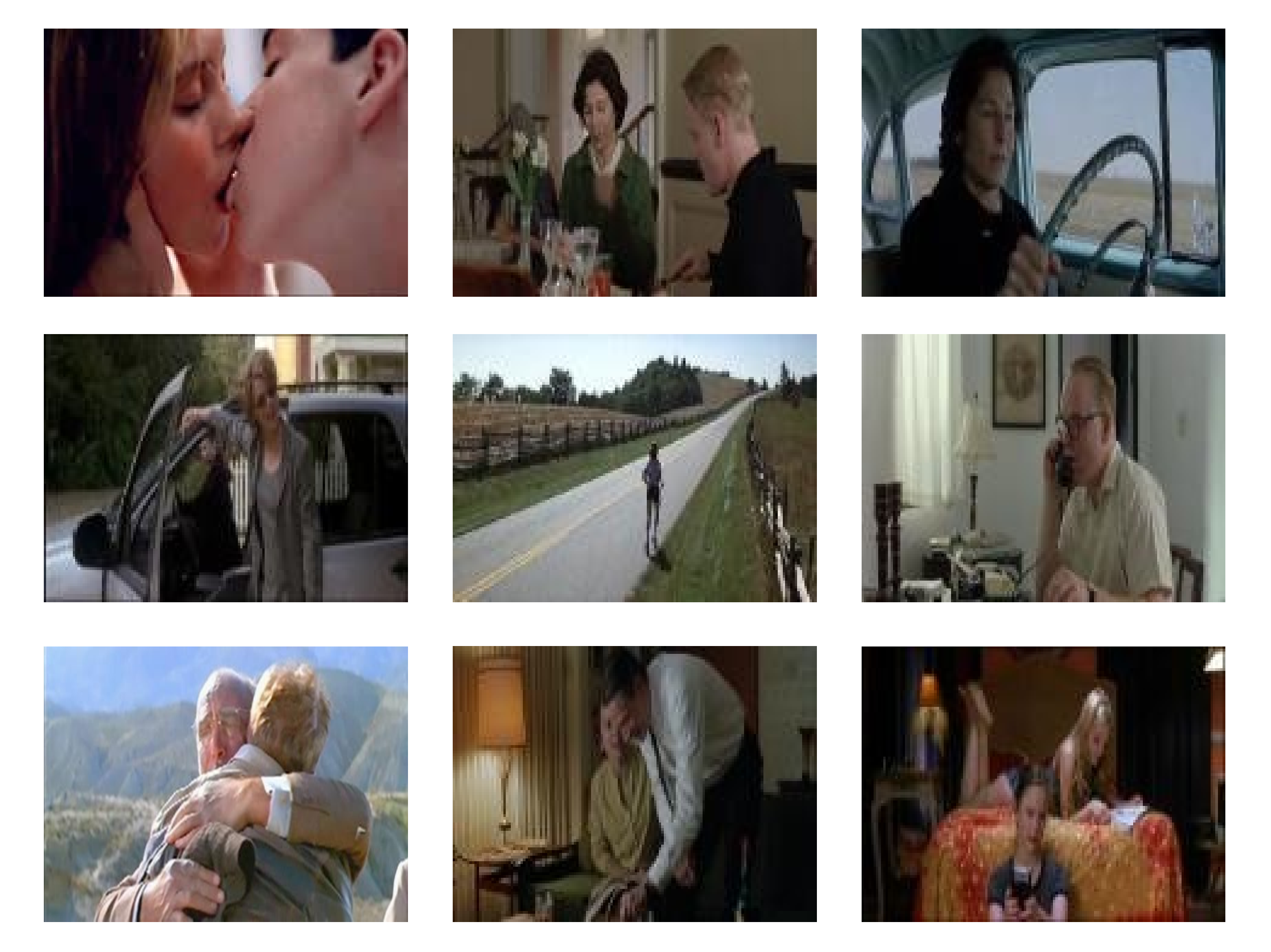,width=8.5cm}}
%\end{minipage}
\caption{Illustration of the Hollywood2 dataset containing human action from Hollywood movies.}%\footnote{Taken from http://www.irisa.fr/vista/actions/hollywood2/}}
\label{fig:res}
\end{figure}

We wanted to evaluate the performance of the descriptors for human action recognition in natural scenarios. Therefore, 
the Hollywood2 dataset~\cite{marszalek09} was a natural choice. This dataset is composed by 12 action classes: answering 
phone, driving car, eating, fighting, getting out of the car, hand shaking, hugging, kissing, running, sitting down, sitting 
up, standing up (see Figure~\ref{fig:res}). Videos were collected from a set of 69 different Hollywood movies, where 33 were used 
to generate the training set and 36 the test set. Action video clips were divided in three separate subsets, namely an automatic 
(noisy) training set, a (clean) training set and the test set. We only used the clean training set containing 823 samples 
and the test set containing 884 samples. 

\subsection{Bag-of-Features Video Representation}
\label{sec:bags}

When spatiotemporal local features are extracted, they only provide a very local and disconnected representation of the video clips. One way to give a more meaningful representation is to use the \emph{bag-of-features} (\textbf{BoF}) approach, which has been successfully applied to many applications of video analysis~\cite{vande10}~\cite{LMSR08}. Using BoF requires the construction of a vocabulary of features (or visual vocabulary). Although this is commonly accomplished by using $k$-means, it is well known that for very high-dimensional spaces, simple clustering algorithms perform badly, and thus a reasonable and efficient choice is just to select a random sample to form the visual vocabulary: this saves computational time and achieves comparable results. The vocabulary size was set to 4000 since this number has empirically demonstrated good results and is consistent with the literature~\cite{LMSR08}~\cite{marszalek09}. At this point, spatiotemporal local features of a video clip is assigned to the closest visual word of the vocabulary (we use Euclidean distance function), which produces a histogram of visual words. This histogram of visual word frequency now accounts for the new video representation.

\subsection{Classification}
\label{sec:class}

To classify the videos, we have used Support Vector Machines (SVM), using the LibSVM~\cite{libsvm} implementation. Since our aim is to highlight the performance of the descriptors, we have chosen to simplify the classifier by using linear kernels (experiments with more complex kernels were performed with comparable results). SVM being a binary classifier, LibSVM implements multi-classification by the one-to-one method, which creates $n(n-1)/2$ binary classifiers (where $n$ is the number of classes) and applies  
a majority voting scheme to assign the class of an unknown element. %This method was proved more efficient than 
%the \emph{one-to-rest} and DAG multi-class methods~\cite{multisvm}.

\subsection{Experimental Protocol}

%The experiment was conducted by following the steps below:

\begin{enumerate}
 \item Extract local features of the whole dataset (using both descriptors, HueSTIP and STIP),
 \item Build the visual vocabularies, one for each feature type (HueSTIP or STIP),
 \item Assemble the histograms of visual words representing each video clips of the dataset,
 \item Learn the classifiers (one for each feature type) of the clean training set using SVM, in which 
       the training and test samples are already separately available, as described in~\ref{sec:dataset},
 \item Classify the samples of the test dataset. 
\end{enumerate}

\subsection{Results and Discussion}

Table~\ref{tab:tab01} evaluates the performance of both descriptors, STIP and HueSTIP, for the human action recognition task. It shows that there exists a gain in using color information for the classification of some actions. Especially, half of the classes had the best performance when using HueSTIP, namely \emph{AnswerPhone}, \emph{FightPerson}, \emph{HugPerson}, \emph{Run}, \emph{SitDown}, and \emph{StandUp}. This increased performance brought by the HueSTIP may have come either from information retrieved from the scene backgrounds or from parts of the objects of interest that is usually ignored by traditional shape/motion descriptors but gains meaning as the color description is considered.

Some assumptions can be made to justify the performance improvements achieved by the HueSTIP at the above actions. For example, for the \emph{AnswerPhone} class, the color information from the background describing the indoor scenario in which this action usually takes place may have added some importance. For the \emph{FightPerson} class, we have that in situations involving aggressive behaviors, the presence of blood can be expected, and it will help define the action if color information is taken into account. Regarding the class \emph{Run}, color information from the outdoor scenario might be useful. However, for many other classes, the addition of color information actually results in losses. This is somewhat intuitive in \emph{DriveCar} and \emph{GetOutCar}, where the color variablity of cars acts more as a confusion than a help. The huge loss in performance in classes like \emph{Eat} and \emph{Kiss}, however was somewhat unexpected and reveal the weakness of using the same neighborhood for extracting the optical flow and color information.

\begin{table}[htb]
\begin{center}
\caption{Performance of the descriptor for each separate action class.}\label{tab:tab01}
\begin{tabular}{|c|c|c|}
\hline \textbf{Action} & \textbf{HueSTIP}& \textbf{STIP}\\
\hline
\hline \emph{AnswerPhone} &\textbf{12.5\%} &9.38\% \\
\hline \emph{DriveCar} &71.57\% &\textbf{76.47\%} \\
\hline \emph{Eat} &45.45\% &\textbf{57.58\%} \\
\hline \emph{FightPerson} &\textbf{68.57\%} &62.86\% \\ 
\hline \emph{GetOutCar} &8.77\% &\textbf{19.3\%} \\ 
\hline \emph{HandShake} &6.67\% &\textbf{8.89\%} \\ 
\hline \emph{HugPerson} &\textbf{18.18\%} &12.12\% \\
\hline \emph{Kiss} &38.83\% &\textbf{49.51\%} \\
\hline \emph{Run} &\textbf{58.87\%} &57.45\% \\
\hline \emph{SitDown} &\textbf{45.37\%} &41.67\% \\
\hline \emph{SitUp} &0.0\% &0.0\% \\
\hline \emph{StandUp} &\textbf{54.11\%} &51.37\% \\
\hline
%\hline \textbf{Average Precision} &35.74\% &\textbf{37.22\%} \\
\hline
\end{tabular}
\end{center}
\end{table}

%\begin{table}[htb]
%\begin{center}
%\caption{General performance comparison.}\label{tab:tab02}
%\begin{tabular}{|c|c|c|}
%\hline \emph{HueSTIP} &\emph{STIP} \\
%\hline
%\hline
%\end{tabular}
%\end{center}
%\end{table}

%\begin{table}[htb]
%\begin{center}
%\caption{General performance comparison.}\label{tab: tab2}
%\begin{tabular}{c|c|c}
%\hline \textbf{RHueSTIP} &\textbf{SHueSTIP} &\textbf{STIP} \\
%\hline
%\hline 35.74 &36.45 &37.22 \\
%\hline
%\end{tabular}
%\end{center}
%\end{table}

\section{Conclusion}
\label{sec:conclusion}

We consider that HueSTIP has showed promising results for a preliminary work: the experiments show it can improve classification rates of actions, but that this improvement tends to be very class dependent. 

We are currently working on some of its interesting issues, especially the reasons why its performance is so unexpectedly low in a few classes. We suspect that using the same feature detector for STIP and HueSTIP might give the former an unfair advantage for the classes where the interesting color phenomena happens at different scales than interesting grayscale phenomena. 

\bibliographystyle{IEEEbib}
\bibliography{strings,refs}

\end{document}